%% file: main.tex
\pdfoutput=1
\documentclass[11pt]{article}

\usepackage[]{acl}
\usepackage{listings}

\usepackage{times}
\usepackage{latexsym}

\usepackage[T1]{fontenc}
\usepackage[utf8]{inputenc}

\usepackage{microtype}
\usepackage{algorithm}
\usepackage{algorithmic}
\usepackage[utf8]{inputenc} % allow utf-8 input
\usepackage[T1]{fontenc}    % use 8-bit T1 fonts
\usepackage{url}            % simple URL typesetting
\usepackage{booktabs}       % professional-quality tables
\usepackage{amsfonts}       % blackboard math symbols
\usepackage{nicefrac}       % compact symbols for 1/2, etc.
\usepackage{microtype}      % microtypography
\usepackage{xcolor}         % colors
\usepackage{enumitem}
\usepackage{multirow}
\usepackage{graphicx}
\usepackage{amsmath}
\usepackage{hyperref}
\hypersetup{
    colorlinks=true,
    linkcolor=black,
    filecolor=magenta,      
    urlcolor=black,
    pdftitle={Overleaf Example},
    pdfpagemode=FullScreen,
    }

\usepackage{subcaption}

\newcommand{\mddial}{{MDDial}}

\title{\mddial: A Multi-turn Differential Diagnosis Dialogue Dataset with Reliability Evaluation}
\author{Srija Macherla \\
  Affiliation / Address line 1 \\
  Affiliation / Address line 2 \\
  Affiliation / Address line 3 \\
  \texttt{email@domain} \\\And
  Second Author \\
  Affiliation / Address line 1 \\
  Affiliation / Address line 2 \\
  Affiliation / Address line 3 \\
  \texttt{email@domain} \\}

  \author{
   Srija Macherla $^1$ \quad Man Luo$^{1,2}$ \quad Mihir Parmar$^1$ \quad \quad Chitta Baral$^1$ \\
    \textsuperscript{1} Arizona State University \quad \textsuperscript{2} Mayo Clinic\\
}

\begin{document}
\maketitle
\begin{abstract}

Dialogue systems for Automatic Differential Diagnosis (ADD) have a wide range of real-life applications.
These dialogue systems are promising for providing easy access and reducing medical costs. 
Building end-to-end ADD dialogue systems requires dialogue training datasets.
However, to the best of our knowledge, there is no publicly available ADD dialogue dataset in English (although non-English datasets exist). 
Driven by this, we introduce \mddial, the first differential diagnosis dialogue dataset in English which can aid to build and evaluate end-to-end ADD dialogue systems. 
Additionally, earlier studies present the accuracy of diagnosis and symptoms either individually or as a combined weighted score. This method overlooks the connection between the symptoms and the diagnosis. 
We introduce a unified score for the ADD system that takes into account the interplay between symptoms and diagnosis. This score also indicates the system's reliability.
To the end, we train two moderate-size of language models on \mddial{}. 
Our experiments suggest that while these language models can perform well on many natural language understanding tasks, including dialogue tasks in the general domain, they struggle to relate relevant symptoms and disease and thus have poor performance on \mddial{}. 
\mddial{} will be released publicly to aid the study of ADD dialogue research.
\footnote{Dataset is available at \url{https://github.com/srijamacherla24/MDDial/tree/main/data}}

\end{abstract}

\input{1_introduction}
\input{related_work}

\input{2_dataset}

\input{3_evaluation}

\input{4_experiments}

\input{5_analysis}

\input{6_conclusion}

\bibliography{anthology,custom}

\clearpage

\appendix

\input{appendix}

\end{document}

%% file: 1_introduction.tex
\section{Introduction}

\begin{figure}[t]
    \centering
    \includegraphics[width=\linewidth]{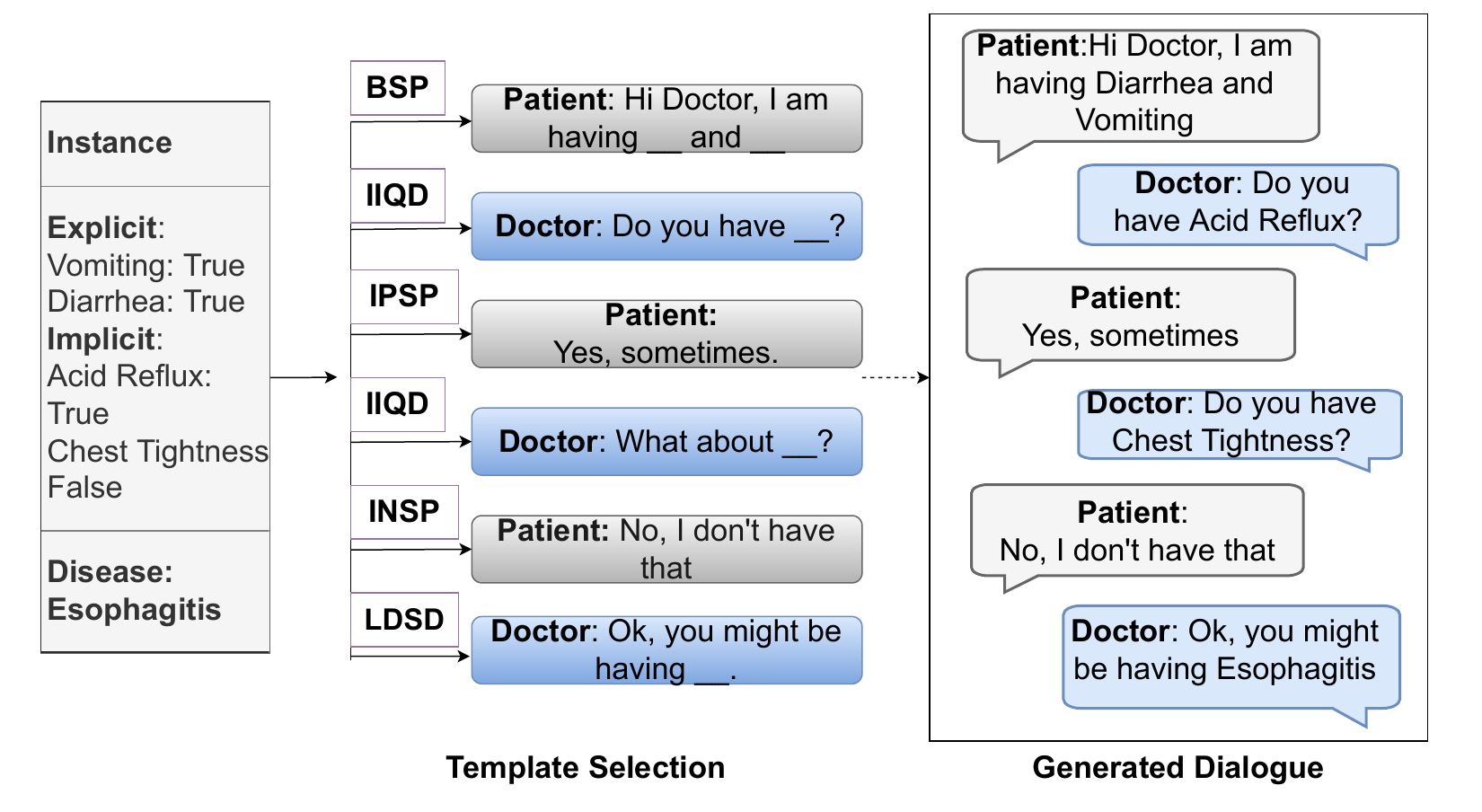}
    \caption{Pipeline of MDDial Generation: Six categories of templates where each has multiple human-curated natural language templates (\S\ref{sec:mddial}).}
    \label{fig:dialogue_generation}
\end{figure}

Automatic Differential Diagnosis (ADD) aims to develop a system that can diagnose patients via interaction with them. 
ADD has many benefits, such as simplifying the diagnostic procedure, reducing the cost of collecting information from patients, and assisting health professionals to make decisions more efficiently~\cite{zeng2020meddialog}. 
To interact with a human, a system needs to understand natural language, which is what recent language models~\cite{devlin2019bert,radford2019language} are good at. 
ADD dialogue datasets are required to either train or evaluate a model.
However, existing English differential diagnosis datasets~\cite{wei2018task, xu2019end, zeng2020meddialog, liu2020meddg, guan2021bayesian} are in a structured format as shown in the left part in Figure \ref{fig:dialogue_generation}.
To the best of our knowledge, there is no publicly available differential diagnosis dialogue dataset in English, such as the right part in Figure \ref{fig:dialogue_generation}. 

Collecting a real-world dialogue dataset is difficult due to the privacy concerns of patients. 
To overcome this issue, we propose a template-based method to create an ADD dialogue dataset based on structured data. 
Specifically, we utilize the Medical Diagnosis Dialogue (MDD) dataset, provided by a competition in ICLR 2020 conference\footnote{https://mlpcp21.github.io/pages/challenge.html}, and define five stages of a differential diagnosis dialogue (\S\ref{sec:mddial}). 
We then create templates for each stage and fill in the templates with explicit symptoms (BSP), implicit symptoms (IIQD), values of each implicit symptom (IPSP and INSP), and the diagnosis (LDSD) given by the MDD dataset. 
We concatenate each sentence to form a complete multi-turn dialogue and name our dataset as \mddial{}.
Figure \ref{fig:dialogue_generation} illustrates the example of a dialogue created from an MDD instance.
Additionally, earlier studies typically assessed individual symptom and diagnosis accuracy, or merely (weighted) summed up two scores. This approach overlooks the connection between symptoms and the resulting diagnosis. 
Logically, a diagnosis should stem from the symptoms, meaning that inquiring about pertinent symptoms should increase the likelihood of an accurate diagnosis compared to unrelated symptoms. Evaluating based on this logic also indicates a system's reliability. A system that produces an accurate diagnosis but inquires about unrelated symptoms is less dependable. Hence, we've introduced a reliability score to gauge a system(\S\ref{sec:metric}). Essentially, a system earns credits when it accurately diagnoses while also pinpointing relevant symptoms.

%%%%% why GPT2 model %%%%%5
% 1. large language models are trained on large corpus, which exhibit world-knowledge and commonsense. 
% 2. they have shown promising results on general domain. 
To the end, we train and evaluate two moderate-size language models, i.e., GPT-2~\cite{radford2019language} and DialoGPT~\cite{zhang2020dialogpt} on \mddial{}, such models can understand natural language in general domains and conduct general tasks, like booking hotel and flights. 
However, our experimental results show that these models struggle with ADD tasks.
Particularly, even models can  
make the correct diagnoses in some cases, they have low-reliability scores on \mddial, and the overall performance has large room to be improved. 
We attribute the low performance to two reasons: (1) the variety of symptoms and diseases make \mddial{} difficult for language models to relate these two. (2) there is not enough training data for the model to learn such a challenging task, which is a common issue in the medical and healthcare domain.
This also suggests that new training paradigms or model architectures are required to transfer general language models to low-resource domains.  
We hope that our dataset can be used to facilitate the development of differential diagnosis in dialogue settings and encourage new models to achieve high reliability. 

%% file: related_work.tex
\section{Related Work}
\label{app:related_work}
\paragraph{Differential Diagnosis}
\citet{wei2018task} proposed task oriented Dialogue System (Muzhi) dataset which has only 4 types of diseases and is in a structured format but not a natural language dialogue format. 
Given the special data structure, they focus only on dialogue management and leave natural language understanding and generation behind. 
\citet{peng2018refuel} proposed reinforcement learning for fast disease diagnosis. \citet{xu2019end} reverse the Muzhi dataset into dialogue form and the resulting dataset DX is smaller. 
However, all these datasets are designed for predicting the pathology a patient is experiencing and do not contain differential diagnosis data \cite{tchango2022ddxplus}.
To facilitate research in diagnosis dialogue systems, large-scale medical dialogue datasets are proposed which include MedDialog \cite{zeng2020meddialog} and MedDG \cite{liu2020meddg}. 
However, these datasets are in the Chinese language. \citet{zeng2020meddialog} created a dataset in Chinese and English language, however, their English version is not publicly available. 
Along with that, the MDD dataset is a similar structure dataset as Muzhi but includes more types of diseases and symptoms. 
Nevertheless, the previous dataset can not be served to train an end-to-end differential diagnosis dialogue system. 
Hence, this work introduces the first English language dataset for building an end-to-end ADD system.
Inspired by previous work \citet{ham2020end}, we train GPT-2 and build an end-to-end dialogue system. 

\paragraph{Template-Based Data Generation} 
Templates have been extensively employed for data generation across various fields~\cite{weston2016towards,banerjee2020can,clark2021transformers,luo2022improving,luo2022biotabqa}.
Using a template-driven generation method can simplify the data production process, particularly in specialized fields where collecting data can be challenging due to its requirement for expert knowledge or sensitivity. 
However, this approach might constrain the range and depth of the resulting data, potentially causing the fine-tuned model to struggle with generalization. 
To mitigate this concern in our study, we test fine-tuned models on two distinct test datasets with varied distributions compared to the training data, ensuring our model doesn't merely adapt to the templates.

%% file: 2_dataset.tex
\section{\mddial} \label{sec:mddial}

\begin{figure}[h]
    \centering
    \includegraphics[width=\linewidth]{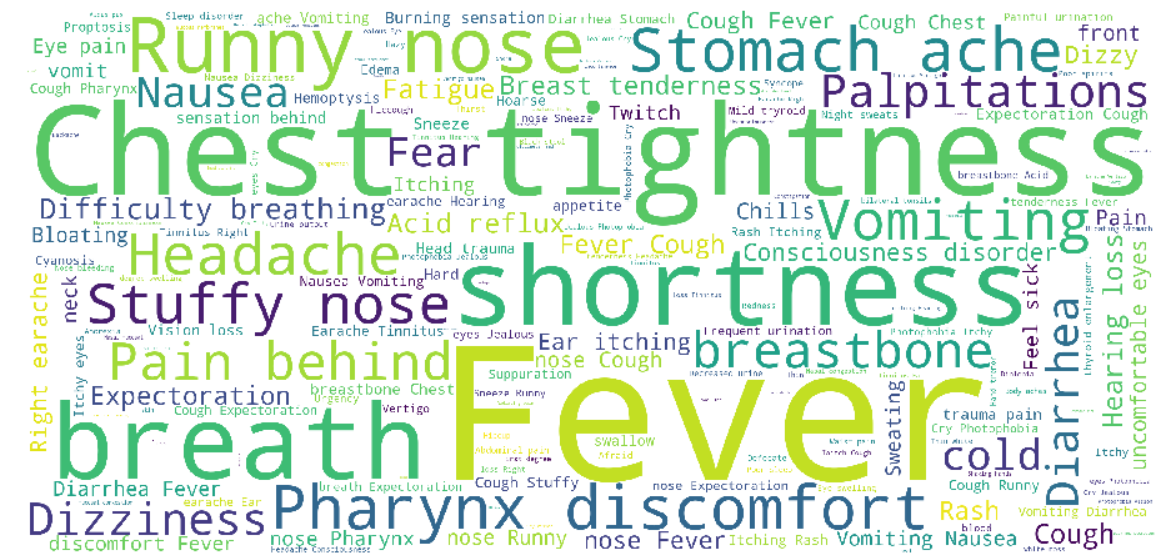}
    \caption{Word cloud indicating diverse symptoms and their frequencies.}
    \label{fig:symptoms_cloud}
\end{figure}

\mddial{} is a template-based dialogue dataset constructed from MDD. 
We choose MDD since it has more variety of diseases and symptoms compared to the previous Muzhi dataset~\cite{wei2018task}. 
MDD includes 12 diseases and 118 symptoms. The dataset has balanced data points about each disease and the frequency of symptoms is given in Figure \ref{fig:symptoms_cloud}.
It shows that \mddial{} has a variety of symptoms with different frequencies which presents a challenge for the model to learn. 

\paragraph{Templates Construction}
% \mddial{} is constructed from different human-authored templates. To generate different templates, 
To simulate a real-life differential diagnosis dialogue, we design five stages that can occur: (1) the Beginning Sentence of a Patient (BSP) which involves explicit symptoms that the patient informs a doctor, (2) Intermediate Inquiry Question of a Doctor (IIQD) which involve implicit symptoms that help the doctor to diagnose the patient, (3) Intermediate Positive answer Sentence of a Patient (IPSP) means that a patient confirms he/she has the inquired symptom, (4) Intermediate Negative answer Sentence of a Patient (INSP) means a patient negates having the inquired symptom, and (5) Last Diagnosis Sentence of a Doctor (LDSD) that informs what kind of disease the patient might have based on the conversation.
For each stage, multiple templates are designed to generate sentences.
The full list of templates is given in Table \ref{tab:templates}.

\input{tables/all_templates}

\paragraph{Quality of Templates} 
First of all, the generation of templates does not necessarily require medical expertise since most people have experience conversing with doctors. 
Given that annotators have commonsense knowledge, we also ask them to study the MedDialogue dataset~\cite{zeng2020meddialog}, this process equips the annotators with domain knowledge if needed\footnote{To avoid confusion, it is worth noticing that the MedDialogue is a 1 to 1 English Dialogue dataset without multi-turn conversation between doctors and patients.}. After the annotators generate the templates, we ask different annotators to verify the fluency, grammar, typographic errors, ambiguity, etc. to guarantee the quality. 

\paragraph{Dialogue Generation} 
Once templates are generated, we use symptoms and the corresponding disease combined with relevant templates to generate dialogue in our dataset. Same templates and procedures have been used to generate training, validation, and testing sets. Statistics of \mddial{} are given in Table \ref{tab:mddial}. 
More examples of dialogues are shown in Appendix \ref{app:dialogue_generation}.

\paragraph{Unseen Templates for Generalization Evaluation}
One potential concern about template-based datasets is that models trained on such texts might only be able to perform effectively on these template sentences.
To mitigate this issue, we design different templates that separate from the training templates to test the generalizability of models.
We use two approaches to obtain the unseen templates. First, we use a T5 model\footnote{\url{https://huggingface.co/ramsrigouthamg/t5_paraphraser}}, and generate multiple paraphrased sentences for given templates\footnote{Paraphrased sentences are checked and updated manually.}. Second, we ask two more annotators to generate new templates without looking at training  templates, and only select those different from the training ones for generalization evaluation. 
% \ref{fig:MDDial_construction} shows schematic representation of dialogue generation in \mddial.

% \begin{figure}[h]
%     \centering
%     \includegraphics[width=.95\linewidth]{images/dialog_datagen_3.png}
%     \caption{Flow chart of dialogue construction and also add example in appendix.}
%     \label{fig:MDDial_construction}
% \end{figure}

% \input{tables/template}

% \paragraph{Dataset Attributes} 

% Statistics of our proposed dataset in Table \ref{tab:mddial} which shows variations in terms of different aspects. \mddial{} provides advantages in two aspects: (1) contains natural and fluent dialogues, and (2) contains diverse set of symptoms. We believe that it is difficult to relate symptoms with the corresponding disease when you have natural language rather than implicit/explicit relations between symptoms and disease is given. This makes \mddial{} a challenging dataset (More details in Appendix \ref{app:dataset_attributes}). 

\input{tables/dataset_statistic}

% \subsection{Review of MDD Dataset} 
% [Srija] in general, describe what is the input and output of the dataset, how does it created and how many diseases, symptoms and dialogue are covered in the dataset. 

%% file: tables/all_templates.tex
\begin{table*}[t]
    \centering 
    \setlength\tabcolsep{4.0pt}
    \footnotesize
    
    \resizebox{\linewidth}{!}{
    \begin{tabular}{c|c|c}
        \toprule
        \textbf{Sentence Type} & \textbf{Templates for Training} & \textbf{Templates for Robustness Testing}
        \\
        \midrule
        \multirow{4}{*}{BSP} & Hi Doctor, I am having \_\_ and \_\_ & Hi Doctor, I feel bad these days, I am suffering from \_\_\\
        \cmidrule{2-3}
        & Recently, I am experiencing \_\_ & Hello Doctor, I am sick and I feel \_\_\\
        \cmidrule{2-3}
        & I have \_\_ and \_\_ & Hey Doctor, thank you for seeing me. I am not well and I have \_\_\\
        \cmidrule{2-3}
        & I have been feeling \_\_ and \_\_ & Hi Doctor, my health is not good, and I feel \_\_\\
        \midrule
        \multirow{4}{*}{IIQD} & Is it? Then do you experience \_\_? & ok, do you feel \_\_ then?\\
        \cmidrule{2-3}
        & In that case, do you have any \_\_? & oh, that is bad, do you have \_\_?\\
        \cmidrule{2-3}
        & What about \_\_? & how about \_\_?\\
        \cmidrule{2-3}
        & Oh, do you have any \_\_? & Tell me, do you feel \_\_?\\
        \midrule
        \multirow{4}{*}{IPSP} & Yes, sometimes. & oh, yes, I forget to mention it. \\
        \cmidrule{2-3}
        & I am experiencing that sometimes. & Yes, I do have. \\
        \cmidrule{2-3}
        & Yes Doctor, I am feeling that as well. & Emm, yes. \\
        \cmidrule{2-3}
        & Yes most of the times. & Yes, I think so. \\
        \midrule
        \multirow{4}{*}{INSP} & No, I don't have that. & It is not my symptom.\\
        \cmidrule{2-3}
        & No, I never had anything like that. & No, I don't think I experienced any. \\
        \cmidrule{2-3}
        & Well not in my knowledge. & I don't think so. \\
        \cmidrule{2-3}
        & Not that I know of. & I don't feel it. \\
        \midrule
        \multirow{4}{*}{LDSD} & In that case, you have \_\_. & Based on your symptoms, you have \_\_.\\
        \cmidrule{2-3}
        & This could probably be \_\_. & From my experience, you can be \_\_.\\
        \cmidrule{2-3}
        & This could probably be \_\_. & You have \_\_.\\
        \cmidrule{2-3}
        & I believe you are having \_\_. & Now I see, you have \_\_.\\
        \bottomrule
        \end{tabular}
        }
\caption{Templates for five types of sentences involved in a differential diagnosis conversation.}
\label{tab:templates}
\end{table*}

%% file: tables/dataset_statistic.tex
\begin{table}[t]
\centering 
\setlength\tabcolsep{4.0pt}
\footnotesize

\resizebox{\linewidth}{!}{
\begin{tabular}{cccc}
\toprule
\textbf{Statistic}  & Train &  Val & Test\\ 
\midrule
Total Dialogues & 1725  &  154 & 237 \\
Avg. turns per dialogue & 5.8 &  7.7 & 5.9 \\
Max. turns in a dialogue & 16 &  15 & 13 \\
Min. turns in a dialogue & 1 &  1 & 1 \\
Avg. words per dialogue & 53.5 &  71.1 & 55.4 \\
Avg. words per patient utterance & 5.6 &  5.6 & 5.6 \\
Avg. words per doctor utterance  & 6.7 &  6.7 & 6.6 \\
\bottomrule
\end{tabular}
}
\caption{Statistics of \mddial{} dataset.}
\label{tab:mddial} 
\end{table}

% \begin{table}[t]
% \centering 
% \setlength\tabcolsep{4.0pt}
% \footnotesize

% \resizebox{\linewidth}{!}{
% \begin{tabular}{cccc}
% \toprule
% \textbf{Statistic}  & Train &  Val & Test\\ 
% \midrule
% Total Dialogues & 1725  &  154 & 237 \\
% Avg. turns per dialogue & 5.8 &  7.7 & 5.9 \\
% Max. turns in a dialogue & 16 &  15 & 13 \\
% Min. turns in a dialogue & 1 &  1 & 1 \\
% Avg. words per dialogue & 53.5 &  71.1 & 55.4 \\
% Avg. words per patient utterance & 5.6 &  5.6 & 5.6 \\
% Avg. words per doctor utterance  & 6.7 &  6.7 & 6.6 \\
% \bottomrule
% \end{tabular}
% }
% \caption{Statistics of \mddial{} dataset.\man{@Srija, could you update this table with the new numbers?}}
% \label{tab:mddial_new} 
% \end{table}

%% file: 3_evaluation.tex
\section{Evaluation} \label{sec:eval}

Firstly, we present metrics to assess an ADD dialogue system. In particular, we design a  disease-wise symptom score to address the limitation of the existing symptom accuracy score. More crucially, we propose a reliability score that not only unifies symptom and disease accuracy but also reflects the degree of reliability of a system. 
Subsequently, we outline an automated evaluation procedure for differential diagnosis. 
While the ideal evaluation procedure will be to ask humans to interact with the model and provide human assessment, this procedure can be expensive and challenging to replicate~\cite{mehri2022report}.
As a result, we design an automated procedure to ensure the conversation's validity and thus mimic the real-life dialogue. 

\subsection{Metrics}

For simplicity, let us assume $\mathcal{D}$ is the set of $n$ number of diseases and $\mathcal{S}$ is the set of $m$ number of symptoms from \mddial{} dataset. Let denote $\mathcal{S}^d$ as subset of symptoms corresponding to disease $d$, where $d \in \mathcal{D}$ and $\mathcal{S}^d \subset \mathcal{S}$. 

\paragraph{Existing Symptom Metric and Issue}\label{sec:metric}
To measure symptom accuracy, \citet{xu2019end} calculates at dialogue-level, denoted as $\mathcal{S}_{score}^{diag}$, which means that a model is credited only if it asks a symptom that exists in the user implicit symptoms for the given dialogue. 
The issue with such an evaluation process is that not all the symptoms in $\mathcal{S}^d$ are present in the given dialogue.
Consequently, even if a model asks for relevant symptoms, it might be punished because of the absence of such symptoms in the ground truth of the implicit symptoms. 
To address this issue, we introduce a disease-wise symptoms score that credits a system if it asks symptoms from $\mathcal{S}^d$.

% To evaluate models, we use our proposed evaluation metrics as well as existing metrics (see .

% Here, we believe that model learns to ask symptoms to diagnose disease $d$ from symptom space $\mathcal{S}^d$ and not merely restricted to pattern of particular dialogue in strict manner. By analyzing the MDD dataset, we find that a disease can be associated with symptoms from $\mathcal{S}^d$, but not all symptoms from $\mathcal{S}^d$ are in every dialogue querying about this disease, thus, even though a inquiry symptom does not appear in the implicit symptoms of a dialogue, it does not mean that this symptom is totally irrelevant. Existing symptom score $\mathcal{S}_{score}^{diag}$ fails to capture this scenario. Moreover, existing metrics merely evaluate model based on either $Diag_{acc}$ or $\mathcal{S}_{score}^{diag}$, however, they fails to analyze relationship between these two scores. When model correctly diagnosis disease, it should ask right symptoms to achieve that and not just predict disease correctly based on its learned disease space $\mathcal{D}$.

\paragraph{Disease-wise Symptom Score} 
% matching score denoted as $\mathcal{S}_{score}^{dise}$ and trustiness score denoted as $\mathcal{T}_{score}$.
$\mathcal{S}_{score}^{dise}$ captures existence of a symptom $s$ for disease $d$ is in $\mathcal{S}^{d}$ and assign score according to Eq. \ref{eq:func_sym_score}.  

\begin{equation}
   f(s, \mathcal{S}^d) = 
    \begin{cases}
      1, & \text{if}\ s \in \mathcal{S}^d \\
      0, & \text{otherwise}
    \end{cases} 
    \label{eq:func_sym_score}
\end{equation}

In addition, not only the quality of the symptoms matter but also the quantities. 
An optimal system should determine the diagnosis by inquiring about the essential symptoms. That is, it shouldn't ask for too few or too many symptoms; otherwise, penalties should be applied.
To this extent, we define a cost function $C$ (Eq. \ref{eq:cost}) based on number of sentences in the original dialogue ($N_{gold}$) and the number of sentences in generated dialogue ($N_{pred}$) to penalize score generated from $f(s, \mathcal{S}^d)$. Together, we define $\mathcal{S}_{score}^{dise}$ as below:

\begin{align}
    C &= \frac{min(N_{gold}, N_{pred})}{max(N_{gold}, N_{pred})}
    \label{eq:cost} \\
    \mathcal{S}_{score}^{dise} &= \frac{C}{N_{pred}} \sum_{\forall s_i \in \mathcal{S}^{diag}} f(s_i, \mathcal{S}^d)
\end{align}

\paragraph{Reliability Score} 
A reliable system should correctly diagnose a patient by asking questions about the related symptoms. In other words, a system is not reliable if it diagnoses correctly but asks for irrelevant symptoms. 
We introduce a reliability score to measure this.
$\mathcal{R}_{score}$ depicts the relationship between $\mathcal{S}_{score}^{dise}$ and the diagnosis accuracy, $Diag_{acc}$. 
Mathematically, 
% Let assume $d_{pred}$ is predicted diagnosis and $d_{gold}$ is ground truth diagnosis for dialogue in evaluation set. 
% For particular dialogue, the $\mathcal{S}_{score}^{dise}$ is higher then some pre-defined threshold ($t$). The intuition behind trustiness score is that we are confident in the model prediction if and only  i.e., $d_{pred}$ and $d_{gold}$ are same. In the case of $\mathcal{S}_{score}^{dise} < t$, we believe that model predict wrong, i.e., $d_{pred}$ and $d_{gold}$ are different. In both the case, model get trustiness score. Now, we define trustiness score as below:

\begin{equation}
   \mathcal{R}_{score}(x) = 
    \begin{cases}
      1, & \text{if}\ \mathcal{S}_{score}^{dise} \geq t \text{ and } Diag_{acc} = 1  \\
      0, & \text{otherwise}
    \end{cases} 
    \label{eq:t_score}
\end{equation}
where $t$ is a threshold of the symptoms scores, a higher value of the threshold means the reliability is more rigid.
% where $x$ is set of $\mathcal{S}_{score}^{dise}, d_{pred}, d_{gold}$ and $t$. To evaluate models, we calculate $\mathcal{T}_{score}$ average over all generated dialogues.

\subsection{Automatic Evaluation Process}

% Previous automatic evaluation work~\cite{}  mainly focus on evaluating how good a sentence that is generated by a system. This is not the major concern in our setting since the quality of the generated sentences can be measured by including the relevant symptoms or the correct diagnosis (symptoms and diagnosis scores introduced in \S\ref{sec:metric}). 

% In our setting, the quality of the generated sentences can be measured by including the relevant symptoms or the correct diagnosis (symptoms and diagnosis scores introduced in \S\ref{sec:metric}). This raise two major issues in automatic evaluation pipeline: (1) \textit{disorder issue}, and (2) \textit{irrelevancy issue}. To address these issues, we design an automatic evaluation pipeline.  

The core difficulty in crafting an automatic evaluation method lies in ensuring dialogue validity.
Given a ground-truth dialogue consisting of a sequence of utterances between a patient and a doctor $\{(p_1, d_1), (p_2, d_2) \dots, (p_n, d_n)\}$, where $p_i$ is a patient utterance, $d_i$ is a doctor utterance. $p_{i+1}$ is inherently a response to $d_i$.
During testing time,  a model initially takes $p_1$ as input and subsequently generates $\hat{d_i}$. The corresponding response $d_i$, which is $p_{i+1}$,  then gets appended to the dialogue sequence  $\{(p_1, \hat{d_1}) \dots, p_i\}$. This sequence serves as input for the model in the next cycle.

A straightforward approach might involve extracting $p_{i+1}$ directly from the ground-truth dialogue to append to the dialogue.  
However, this presents a complication: if $\hat{d_i}$ misaligns with the symptoms described in $d_i$, then $p_{i+1}$ could be an inappropriate response to $\hat{d_i}$.
To address this \textit{disorder issue}, we do not use the ground truth patient utterances in the testing set. Instead, we assess the symptoms mentioned in $\hat{d_i}$ using the ground truth of these symptoms, we then determine the valid $\hat{p_{i+1}}$. If the query asks for unrelated symptoms, we opt for a negative answer sentence as $\hat{p_{i+1}}$.

% \paragraph{Irrelevancy Issue} A system might make mistakes and ask about irrelevant symptoms in the dialogue. To deal with this \textit{irrelevance issue}, we always use negative response (INSP) to make the dialogue continue. 

%% file: 4_experiments.tex
\section{Experimental Setup}\label{sec:experiment}

\paragraph{Models}
We conduct all experiments using the GPT-2 and DialoGPT models. These models have demonstrated their effectiveness in general dialogue tasks~\cite{budzianowski2019hello}. Notably, we opt for a moderately-sized model instead of larger ones like GPT-3. This decision is driven by the sensitive nature of the biomedical and clinical domains, where it is essential to deploy a fine-tuned local model.
In addition, established by 
previous work~\cite{gururangan2020don,luo2022improving} that adapting models to the relevant domain is crucial, however, both GPT-2 and DialoGPT models are not pretrained in the biomedical domain. Therefore, we pre-train both models on MedDialogue~\cite{zeng2020meddialog} in which each dialogue has one utterance from the patient and doctor, respectively. 
We process the input data following the setting as in \cite{zeng2020meddialog}, and the input format is described in the following. 
\paragraph{Input format} We process the training process similar to~\cite{zeng2020meddialog}. In detail, given a dialogue containing a sequence of alternating utterances between patient and doctor, we convert it into a set of pairs \{($s_i$; $t_i$)\} where
the target $t_i$ is a response from the doctor and the source $s_i$ is the concatenation of all utterances (from both patient and doctor) before $t_i$ in a particular dialogue. A dialogue generation model takes $s_i$ as input and generates $t_i$. To distinguish the patient and doctor sentence, we add special words ``patient: '' in front of each patient utterance and ``Doctor: '' in front of each doctor utterance. We set the maximum input length as 512 and add special tokens <pad> to the GPT-2 vocabulary.

\paragraph{Training and Generation Details} 
The models are trained with a learning rate of $5e^{-4}$ for 10 epochs with AdamW Optimization of $1e^{-8}$ as epsilon. The block size used is 512 with a batch size of 2 for both training and evaluation. During the evaluation, the greedy search decoding method is used with a non-repeat ngrams of size 3, and temperature 0.8 to generate dialogues by the doctor.

\paragraph{Two Evaluation Settings} We fine-tune both pre-trained models on \mddial. To evaluate models, we create two evaluation sets: (1) contains a dialogue with seen templates which has a similar distribution as training, and (2) contains a dialogue with unseen templates which has different distribution as training. 
The second evaluation set is created to analyze the generalizability of models. We evaluate models using metrics discussed in Section \ref{sec:metric}. We compute $\mathcal{R}_{score}$ for all possible thresholds ($t$).

% By using different ways of transfer learning, we experiment with four different models,

% A general description of GPT2 and the training objective equations. 

% \paragraph{Transfer Learning}  
% [Shall we use phrase transfer learning on pretraining?]
% Since our dataset is relative small compared to other dataset, and many previous work have shown that in low-resource domain, pretraining is helpful to transfer to target domain. Inspired by this, we test two types of transfer learning, one is transferring the knowledge from the general dialogue domain. To do so, we fine-tune DialoGPT on our dataset. 
% The second one is transferring the knowledge from the biomedical domain dialogue. We use MedDialogue  dataset~\cite{zeng2020meddialog}, in which each dialogue has one utterance from patient and doctor respectively. 

% \paragraph{Training Details}

% \paragraph{Evaluation on Human-curated Data} 

% All of our three, write down some natural language sentence for 10 dialogues. (only for patients) 

%% file: 5_analysis.tex
% \newpage
\section{Results and Analysis}

Table \ref{tab:main_result} shows the performance of models on three evaluation scores on evaluation set (1) and the average number of sentences per dialogue generated by the models. 
From Table \ref{tab:main_result}, we can observe that DialoGPT outperforms GPT-2 in terms of $\mathcal{S}_{score}^{dise}$ and $Diag_{acc}$. In addition, DialoGPT generates fewer sentences per dialogue on average compared to GPT-2 which indicates that DialoGPT can diagnose the patient faster and more accurately than GPT-2. 
% DialoGPT is pre-trained on a large dialogue dataset from the general domain. 
This result implies that pretraining on the general dialogue dataset allows the model to acquire more understanding of dialogue and indeed help in improving model performance. 
Moreover, we observe that our
$\mathcal{S}_{score}^{dise}$ is more coherent with $Diag_{acc}$ of two models compared to $\mathcal{S}_{score}^{diag}$ (i.e. a system has a higher diagnosis score also has a higher symptom score). 
This shows that our disease-wise symptoms score is a better metric to evaluate the model than the dialogue-wise symptoms score.

% While the focus of this paper is not to propose a better ADD system, we conduct error analysis of the prediction of the DialoGPT to understand the current model, which can provide guidance to build future ADD systems (see analysis in Appendix \ref{apd:analysis}). 

\input{tables/main_result}

\paragraph{Reliability} 
Figure \ref{fig:reliability} shows the trend of $\mathcal{R}_{score}$ for both models (detailed numbers are given in Appendix \ref{app:reliability}).
Observing Figure \ref{fig:reliability}, it's evident that at even a moderate threshold (for instance, 0.5), the reliability scores for both models are below 0.2. 
This outcome underscores that, even when the models diagnose accurately, they don't necessarily draw from relevant symptoms. More crucially, this suggests that the diagnosis score doesn't truly capture the extent of reliability. 
Additionally, as the threshold rises, DialoGPT's reliability sees a more significant impact compared to GPT-2. At higher threshold values, both models exhibit comparable reliability, indicating that DialoGPT doesn't necessarily outshine GPT-2 in terms of reliability.

\begin{figure}[ht]
    \centering
    \includegraphics[width=0.9\linewidth]{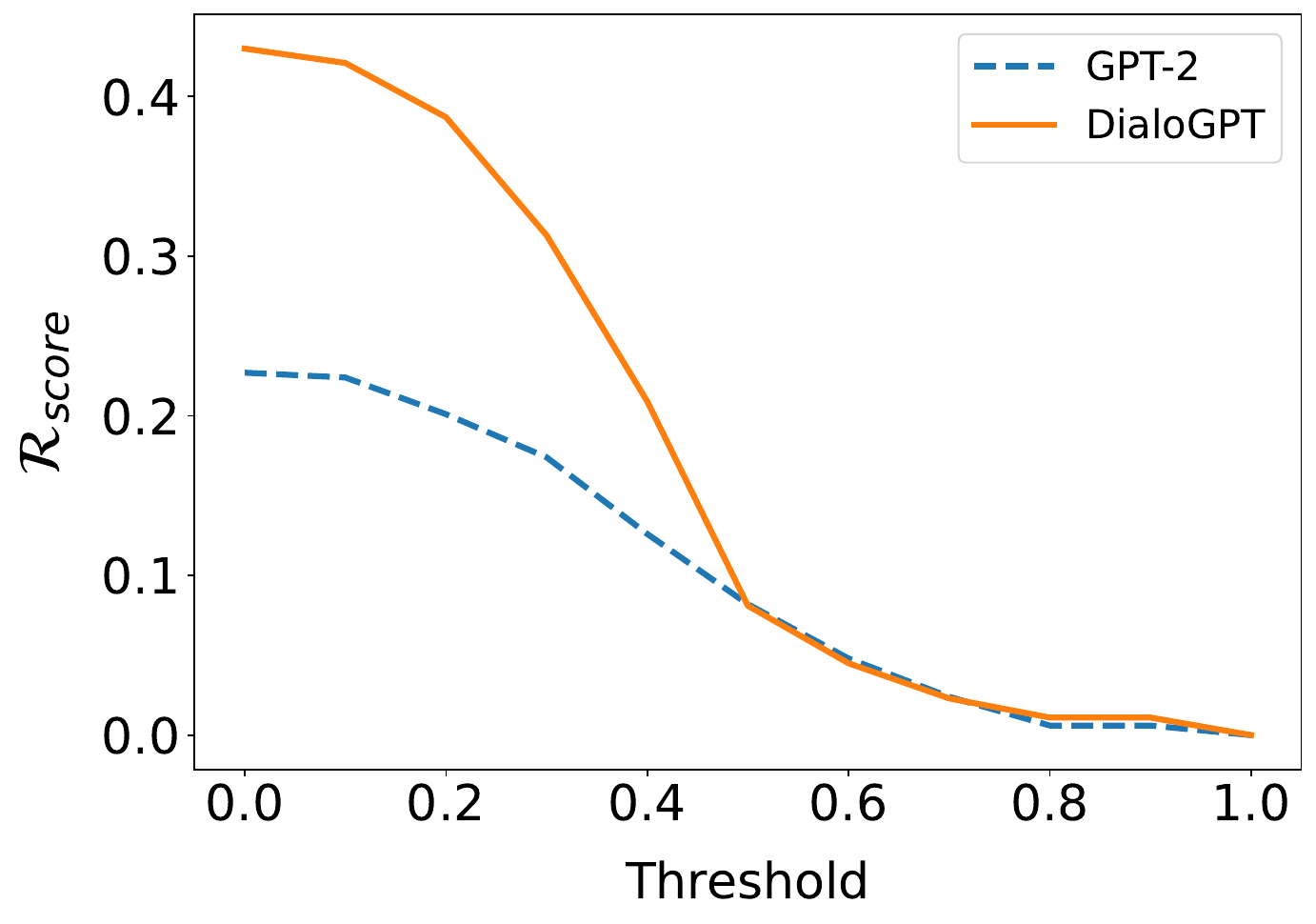}
    \caption{$\mathcal{R}_{score}$ for GPT-2 and DialoGPT on all possible thresholds.}
    \label{fig:reliability}
\end{figure}

\begin{figure}[ht]
\noindent\begin{subfigure}[b]{0.25\textwidth}
    \includegraphics[width=0.99\textwidth]{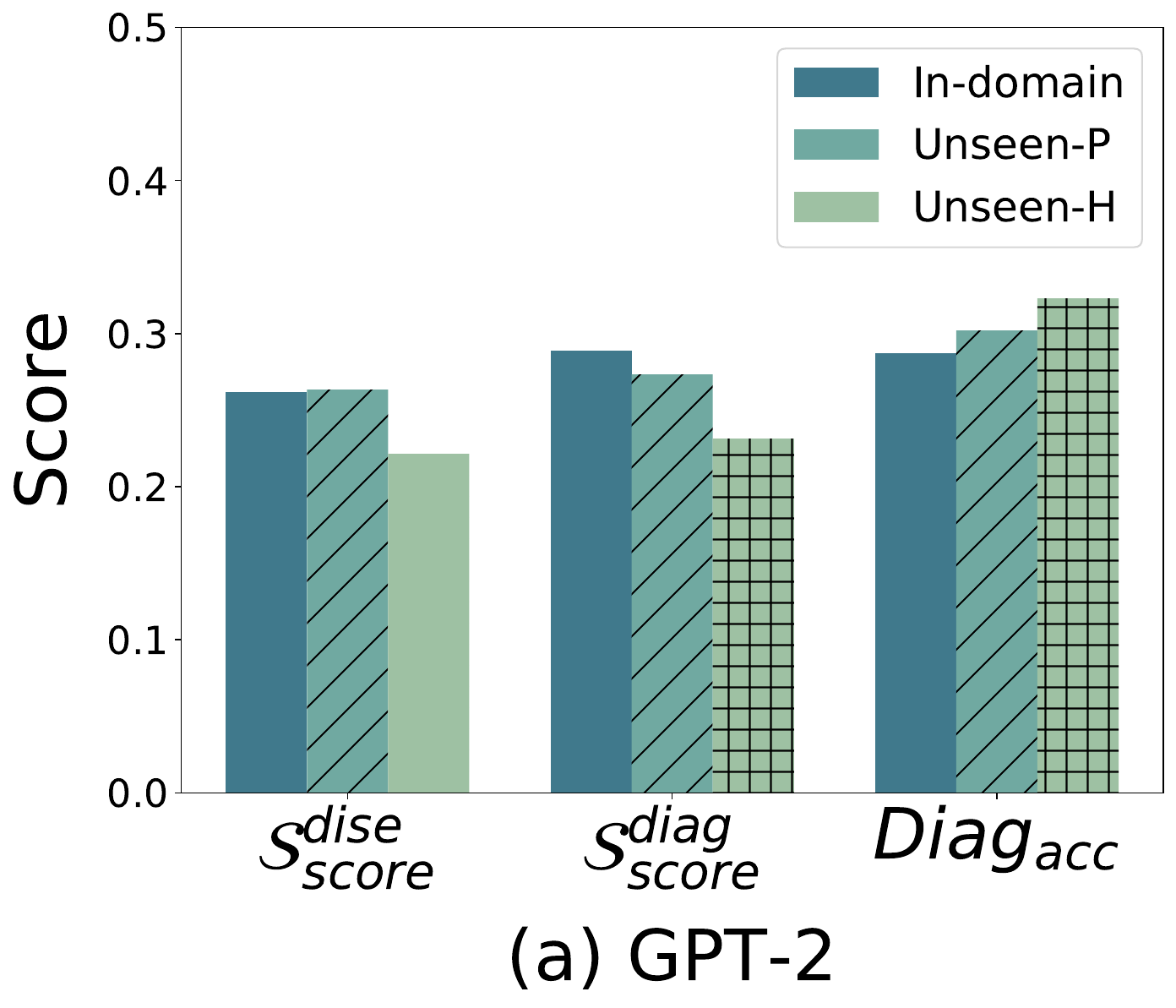}
\end{subfigure}%
\noindent\begin{subfigure}[b]{0.25\textwidth}
    \includegraphics[width=0.99\textwidth]{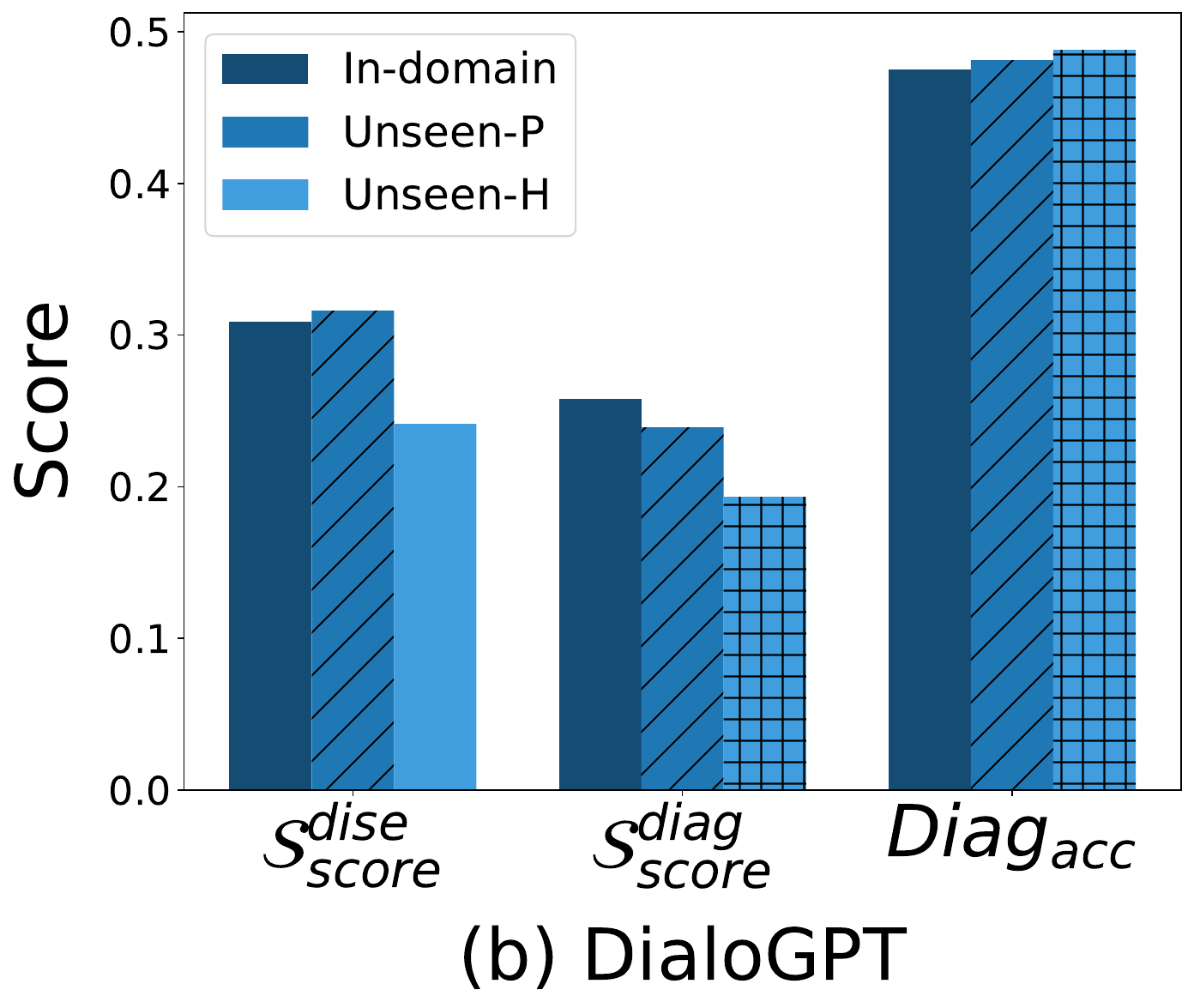}
\end{subfigure}%
\caption{
Performance of two models on three datasets: in-domain, paraphrased dataset (Unseen-P), and other author hand-written templates dataset (Unseen-H).}
\label{fig:three_datasets}
\end{figure}

\paragraph{Generalizability} We evaluate both models on dialogues generated with unseen templates to address that model will not heavily rely on the templates and show generalization.  
From the Figure~\ref{fig:three_datasets}  (detailed numbers are given in Appendix \ref{app:generalization}), we have three observations. First, the performance of the paraphrasing templates is similar to the in-domain testing set \ref{tab:main_result}.
Second, the human-authored templates with different distributions have almost similar $Diag_{acc}$ for both models compared to in-domain templates, and this results in lower trustiness score. 
Third, both models achieve better diagnosis performance on unseen templates but lower performance on the symptom scores. 
In summary, the overall performances of two models on unseen templates are comparable with evaluation on in-domain templates but with lower trustiness scores.

\input{error_analysis}

% \subsection{Analysis}

% Conduct Error Analysis, which disease the model are more likely to failed, which symptoms the models are used to ignore.
% the accuracy for the first sentence and the other sentence.

% \begin{itemize}
% \item why the higher the diagnosis scores is, the lower symptoms score is? (1) Let's change the way how to calculate the symptom score. for each disease, get all the associate symptoms from the training and testing data, then calculate the score. Hopefully if the diagnosis score is higher, the symptom score is higher; otherwise, lower.  If this happens, then recalculate the trust score for incorrect and correct dialogue (the current ones do not make sense). 
% (2) if (1) does not work, let's check the average sentences of correct dialogue and incorrect dialogues.  

% \item calculate the accuracy respect to each disease. 
% also see the percentage of dialogues of each disease, is the accuracy related to the percentage. 

% \item pretrained on the meddialogue for the GPT2 is helpful but it is not helpful for the dialogGPT2, why? For this one, compare the prediction of row4 and row3, in what cases, row4 predict correct but row3 does not; in what cases, row3 predict correct but row4 does not. Similarly, compare row1 and row2. 
% \end{itemize}

%% file: tables/main_result.tex
\begin{table}[ht]
    \centering 
    \setlength\tabcolsep{3.0pt}
    \footnotesize
    
    \resizebox{\linewidth}{!}{
    \begin{tabular}{ccccc}
        \toprule
        Model & $\mathcal{S}_{score}^{diag}$ & $\mathcal{S}_{score}^{dise}$ & $Diag_{acc}$ & Avg. Sentence
        \\
        \toprule
        GPT-2
        & \textbf{0.289}
        & 0.262
        & 0.287
        & 7.64 \\
        DialoGPT
        & 0.258
        & \textbf{0.309}
        & \textbf{0.475}
        & 6.13 \\
        \bottomrule
        \end{tabular}
        }
\caption{Performance of both models on \mddial.}
\label{tab:main_result}
\end{table}

%% file: error_analysis.tex
\section{Error Analysis} \label{apd:analysis}

While the focus of this paper is not to propose a better ADD system, we conduct an error analysis of the prediction of the DialoGPT to understand the current model, which can provide guidance to build future ADD systems.
We conduct the following analysis based on our best model’s (DialoGPT) prediction, we answer two questions below. 

\subsection{What diseases are more difficult to predict?}

There are 12 diseases and for each disease $D$, we compute the individual disease accuracy, i.e. the ratio of dialogues predicted correctly by the model. 
The accuracy of each disease is listed in Table \ref{tab:disease_acc}, the least accuracy means the more difficult for the model to predict and the three most difficult diseases for the model are Coronary heart disease, Traumatic brain injury, and External otitis.

\begin{table}[t]
\centering 
\setlength\tabcolsep{4.0pt}
\footnotesize

\resizebox{0.8\linewidth}{!}
{
\begin{tabular}{cc}
\toprule
\textbf{Disease}  & \textbf{Accuracy} \\ 
\toprule
Coronary heart disease & 0.0 \\
External otitis &  0.0 \\
Traumatic brain injury & 0.0 \\
Mastitis & 0.27 \\
Pneumonia & 0.40 \\
Asthma & 0.53 \\
Esophagitis & 0.56 \\
Rhinitis & 0.60 \\
Thyroiditis & 0.62 \\
Enteritis & 0.71 \\
Conjunctivitis & 0.76 \\
Dermatitis & 1.0  \\
\bottomrule
\end{tabular}
}
\caption{The individual disease accuracy.}
\label{tab:disease_acc} 
\end{table}

\subsection{Are there any symptoms ignored by the dialogue systems?}
To answer this question, for each symptom $S$, we compute the frequency of $S$ in the testing set ($f_1$), and the frequency of $S$ in the prediction of model ($f_2$). 
% and we sort the symptoms based on $\frac{f_2}{f_1}$. 
The lower value of $\frac{f_2}{f_1}$, the more easily the symptom is ignored by the model. 
For simplicity, Table \ref{tab:symptom_acc} lists 10 symptoms which have the lowest value of $\frac{f_2}{f_1}$ among all symptoms.   

\begin{table}[t]
\centering 
\setlength\tabcolsep{4.0pt}
\footnotesize

\resizebox{\linewidth}{!}
{
\begin{tabular}{cc}
\toprule
\textbf{Symtom}  & $\frac{f_2}{f_1}$ \\ 
\toprule
tinnitus & 0.22 \\
rash & 0.27 \\
pain in front of neck & 0.35 \\
itchy eyes & 0.39 \\ 
chest tightness and shortness of breath & 0.45 \\ 
itching & 0.48 \\
suppuration & 0.50 \\ 
difficulty breathing & 0.52 \\
twitch & 0.52 \\
sneeze & 0.54 \\
mild thyroid enlargement & 0.58 \\
\bottomrule
\end{tabular}
}
\caption{The 10 most difficult symptoms for the model.}
\label{tab:symptom_acc} 
\end{table}

%% file: 6_conclusion.tex
\section{Conclusions}

This paper introduced \mddial, the first publicly available dataset for training ADD systems in the English language. With this, we proposed a new evaluation technique, i.e., trustiness to evaluate existing ADD systems in terms of symptom score and diagnosis accuracy. We evaluate existing state-of-the-art models in the dialogue domain, GPT-2 and DialoGPT on \mddial. Experimental results show that these models are not able to relate symptoms with their corresponding disease in \mddial. Our analysis shows that ADD task becomes difficult for the model when dialogues are in natural language and contain diverse symptoms. This suggests that \mddial{} is a challenging dataset and indicates future research scope in improving ADD systems. 

\section*{Limitations}

There are a couple of improvements we intend to do in the future. 
First, \mddial{} is generated based on templates, the diversity of the language is limited, and also the complexity of the language might be too simple to compare to the real-world dialogue between doctors and patients~\cite{maynard1991interaction,asp2010language}.  
A potential solution will be to utilize an LLM to increase both the diversity and the complexity.
Second, we use moderate-size language models and test them on \mddial{}. To see the full ability of the language models, we can adapt the prompting or multi-modular methods as proposed in \cite{hosseini2020simple, lee2021dialogue}. 
Large language models like ChatGPT, known for their proficiency in understanding dialogue, raise the intriguing question of whether they can effectively handle ADD tasks. 
While these models might face hurdles due to limited biomedical domain expertise, in-context learning~\cite{hu2022context,parmar2022boxbart,luo2023dr} offers a promising avenue to address this issue.
% Second, the symptom score and diagnosis score are computed based on exact matching. Nevertheless, normalization is important for biomedical terminology~\cite{tsuruoka2008normalizing}, and thus, using methods such as UMLS~\cite{bodenreider2004unified} to normalize the predicted symptoms might improve models' performance.
Third, while human evaluation itself has drawbacks, it can reveal how well a differential diagnosis system interacts with humans. Thus, we plan to include human evaluation in the future to deeply analyze the model performance. 

%% file: appendix.tex
\section{Dialogue Generation}
\label{app:dialogue_generation}

% \begin{figure}[ht]
%     \centering
%     \includegraphics[width=\linewidth]{images/block_diag.png}
%     \caption{Flow Diagram of Dialogue Generation}
%     \label{fig:dialogue_generation}
% \end{figure}

Table \ref{tab:templates} presents different templates used for generating the dialog Dataset from the MDD dataset for training and testing the DialoGPT and GPT-2 models. Each sentence type has 4 templates from which a template is chosen randomly based on the dialogue's sequence, i.e., if a dialogue is in its initial phase, then the patient's initial sentence is chosen from the Beginning Sentence of a Patient (BSP) template set. Similarly, the next expected dialogue would be the doctor asking if the patient is experiencing any other symptoms. For this scenario, a template is fetched from the Intermediate Inquiry Question of a Doctor (IIQD) template set. If the patient is experiencing the same, a positive reply from the patient is expected, which is chosen from  Intermediate Positive answer Sentence of a Patient (IPSP) template set, or for a negative response, a template is selected randomly from the Intermediate Negative answer Sentence of a Patient (INSP) template set. Finally, the Last Diagnosis Sentence of a Doctor (LDSD) template set is used for the final sentence which is the diagnosis provided by the doctor.

Figure \ref{app:dialogue_generation} depicts the flow diagram of dialogue generation. An instance from the MDD dataset as shown in figure 4 consists of explicit symptoms which the patient informs the doctor, implicit symptoms that the doctor draws the information from the patient based on the patient's input, and the disease that the doctor diagnose. The template selection model converts this information to patient's and doctor's utterances as described above using the templates from Table 6. The third step depicts the randomly chosen templates from their respective template sets as per the sequence of the dialogue. The placeholders for symptoms and diseases are replaced once the template is selected, finally, a dialogue is generated with multiple patient's and doctor's utterances as shown in the last step of the diagram. 

\input{tables/reliability_scores}

\section{Dialog Examples}
\label{app:dialog_examples}

\textbf{Example 1}

\noindent Patient: I have been feeling Chills and fever.

\noindent Doctor: Oh, do you have any Breast tenderness?

\noindent Patient: Yes most of the times.

\noindent Doctor: What about Body aches?

\noindent Patient: Yes Doctor, I am feeling that as well.

\noindent Doctor: What about Fatigue?

\noindent Patient: Yes, sometimes.

\noindent Doctor: This could probably be Mastitis.

\vspace{3mm}

\noindent \textbf{Example 2}

\noindent Patient: Hi Doctor, I am having Pharynx discomfort and Stuffy nose.

\noindent Doctor: What about Cough?

\noindent Patient: No, I don't have that.

\noindent Doctor: In that case, do you have any Fever?

\noindent Patient: Well not in my knowledge.

\noindent Doctor: Oh, do you have any Ulcer?

\noindent Patient: No, I don't have that.
\noindent Doctor: Is it? Then do you experience First degree                swelling of bilateral tonsils?

\noindent Patient: Yes Doctor, I am feeling that as well.

\noindent Doctor: I believe you are having Rhinitis.

% \section{Dataset Attributes}
% \label{app:dataset_attributes}

% \mddial{} dataset is the first publicly available English language dataset to train models for ADD tasks. We utilize all diseases and symptoms from the MDD dataset. \mddial{} dataset spans over 12 diseases and 118 symptoms in total. Furthermore, we provide different statistics of our proposed dataset in Table \ref{tab:mddial} which shows variations in terms of different aspects.

% \paragraph{Naturalness and Fluency} We design a templates (see Table \ref{tab:templates}) more closer to real-life scenarios which makes dialogues in \mddial{} more natural and fluent. This can be advantageous and challenging while training ADD systems. We believe that it is difficult to relate symptoms with the corresponding disease when you have natural language rather than implicit/explicit relations between symptoms and disease is given.

% \section{Existing Metrics}
% \label{app:existing_metrics}

% Similar to \citet{wei2018task}, we use the rate of making the right diagnosis as dialogue accuracy denoted as $Diag_{acc}$. For symptoms evaluation, we use a symptom matching rate from \citet{xu2019end}. This score is calculated at dialogue-level denoted as $\mathcal{S}_{score}^{diag}$ where a successful matching means models ask a symptom that exists in user implicit symptoms for a particular dialogue. Along with that, we also use standard evaluation F1, Recall, and Accuracy score to measure the symptom matching score~\cite{zeng2020meddialog}.

% \input{error_analysis}

\section{Reliability Evaluation} \label{app:reliability}

Table \ref{tab:reliability} shows the details number of $\mathcal{T}_{score}$ presented in Figure \ref{fig:reliability}. This $\mathcal{T}_{score}$ is across all possible thresholds ranging between [0,1] with 0.1 step.

\section{Generalizability} \label{app:generalization}

Table \ref{tab:robustness_1} and Table \ref{tab:robustness_2} show the scores corresponding to Figure \ref{fig:three_datasets}.

\input{tables/robustness}

%% file: tables/reliability_scores.tex
\begin{table*}[t]
    \centering 
    \setlength\tabcolsep{3.0pt}
    \footnotesize
    
    \resizebox{0.85\linewidth}{!}{
    \begin{tabular}{ccccccccccc}
        \toprule
        Model & 0 & 0.1 & 0.2 & 0.3 & 0.4 & 0.5 & 0.6 & 0.7 & 0.8 & 0.9
        \\
        \toprule
        GPT-2
        & 0.227
        & 0.223
        & 0.207
        & 0.173
        & 0.127
        & 0.082
        & 0.047
        & 0.024
        & 0.007
        & 0.007 \\
        DialoGPT
        & 0.43
        & 0.421
        & 0.383
        & 0.309
        & 0.209
        & 0.081
        & 0.048
        & 0.02
        & 0.011
        & 0.011 \\
        \bottomrule
        \end{tabular}
        }
\caption{Trustiness scores of both models on \mddial{} with different thresholds.}
\label{tab:reliability}
\end{table*}

%% file: tables/robustness.tex
\begin{table}[ht]
    \centering 
    \setlength\tabcolsep{4.0pt}
    \footnotesize
    
    \resizebox{0.8\linewidth}{!}{
    \begin{tabular}{cccc}
        \toprule
        \textbf{Model} & \textbf{$\mathcal{S}_{score}^{dise}$} & \textbf{$\mathcal{S}_{score}^{diag}$} & \textbf{$Diag_{acc}$} 
        \\
        \toprule
        GPT-2
        & 0.263
        & 0.273
        & 0.302 \\
        DialogGPT
        & 0.316
        & 0.239
        & 0.481 \\
        \bottomrule
        \end{tabular}
        }
\caption{Performance of both models on the paraphrased evaluation set.}
\label{tab:robustness_1}
\end{table}

\begin{table}[ht]
    \centering 
    \setlength\tabcolsep{4.0pt}
    \footnotesize
    
    \resizebox{0.8\linewidth}{!}{
    \begin{tabular}{cccc}
        \toprule
        \textbf{Model} & \textbf{$\mathcal{S}_{score}^{dise}$} & \textbf{$\mathcal{S}_{score}^{diag}$} & \textbf{$Diag_{acc}$}
        \\
        \toprule
        GPT-2
        & 0.221
        & 0.231
        & 0.323 \\
        DialogGPT
        & 0.241
        & 0.193
        & 0.488 \\
        \bottomrule
        \end{tabular}
        }
\caption{Performance of both models on the human-authored unseen templates evaluation set.}
\label{tab:robustness_2}
\end{table}